\documentclass[sigconf]{acmart}

\AtBeginDocument{%
  }

\copyrightyear{2023}
\acmYear{2023}
\setcopyright{acmlicensed}
\acmConference[MM '23] {Proceedings of the 31st ACM International Conference on Multimedia}{October 29--November 3, 2023}{Ottawa, ON, Canada.}
\acmBooktitle{Proceedings of the 31st ACM International Conference on Multimedia (MM '23), October 29--November 3, 2023, Ottawa, ON, Canada}
\acmPrice{15.00}
\acmISBN{979-8-4007-0108-5/23/10}
\acmDOI{10.1145/3581783.3612025}
\usepackage{balance}
\begin{document}
\title{Distortion-aware Transformer in 360° Salient Object Detection}

\author{Yinjie Zhao}
\affiliation{%
  \institution{Beihang University}
  \city{Beijing}
  \country{China}}
  \email{yjzhao1998@buaa.edu.cn}
\author{Lichen Zhao}
\affiliation{%
  \institution{Beihang University}
  \city{Beijing}
  \country{China}}
  \email{zlc1114@buaa.edu.cn}
\author{Qian Yu$^\ast$}
\affiliation{%
  \institution{Beihang University}
  \city{Beijing}
  \country{China}}
  \email{qianyu@buaa.edu.cn}
\author{Lu Sheng}
\affiliation{%
  \institution{Beihang University}
  \city{Beijing}
  \country{China}}
  \email{lsheng@buaa.edu.cn}
\author{Jing Zhang}
\affiliation{%
  \institution{Beihang University}
  \city{Beijing}
  \country{China}}
  \email{zhang_jing@buaa.edu.cn}
\author{Dong Xu}
\affiliation{%
  \institution{The University of Hong Kong}
  \city{Hong Kong}
  \country{China}}
  \email{dongxu@hku.hk}

\renewcommand{\shortauthors}{Zhao et al.}




\begin{abstract}
With the emergence of VR and AR, 360° data attracts increasing attention from the computer vision and multimedia communities. Typically, 360° data is projected into 2D ERP (equirectangular projection) images for feature extraction. However, existing methods cannot handle the distortions that result from the projection, hindering the development of 360-data-based tasks. Therefore, in this paper, we propose a Transformer-based model called DATFormer to address the distortion problem. We tackle this issue from two perspectives. 
Firstly, we introduce two distortion-adaptive modules. The first is a Distortion Mapping Module, which guides the model to pre-adapt to distorted features globally. The second module is a Distortion-Adaptive Attention Block that reduces local distortions on multi-scale features. 
Secondly, to exploit the unique characteristics of 360° data, we present a learnable relation matrix and use it as part of the positional embedding to further improve  performance. Extensive experiments are conducted on three public datasets, and the results show that our model outperforms existing 2D SOD (salient object detection) and 360 SOD methods.
The source code is available at \href{https://github.com/yjzhao19981027/DATFormer/}{https://github.com/yjzhao19981027/DATFormer/}.
\end{abstract}

\ccsdesc[500]{Computing methodologies~Interest point and salient region detections}
\ccsdesc[500]{Computing methodologies}

\keywords{360 SOD, transformer, distortion-adaptive, positional embedding}

\maketitle

\let \mybackup \thefootnote
\let \thefootnote \relax
\footnotetext{$^*$Qian Yu is the corresponding author.}
\let \thefootnote \mybackup
\let \mybackup \imareallyundefinedcommand

\section{Introduction}
\begin{figure}[t]
\centering
\includegraphics[width=1\linewidth]{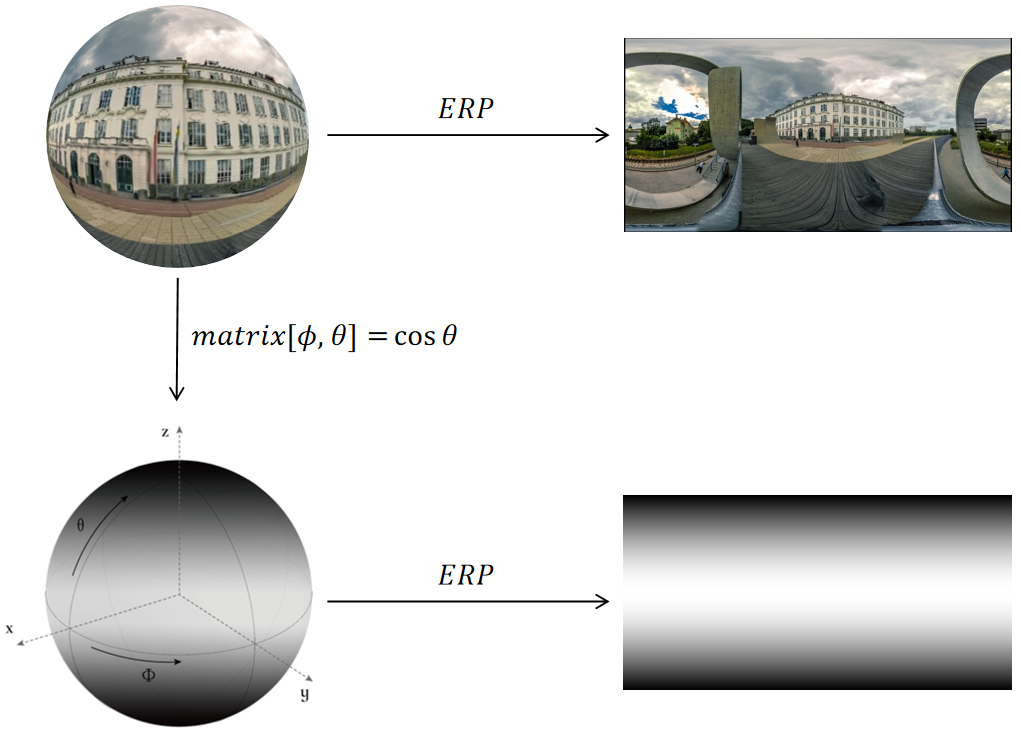}
\caption{\textbf{Top}: The 360° data is mapped to an ERP image via equirectangular projection, during which the geometric deformation is introduced. \textbf{Bottom}:
The Motivation of Relation Matrix. In most 360° images, the most salient objects are located near the equator, which corresponds to the middle region in an ERP image. In this work, we use the cosine function to encode the spatial prior on the sphere.}
\label{fig:motivation}
\end{figure}

In recent years, with the development of virtual reality (VR) and augmented reality (AR) technologies, a new form of data called \textit{360° omnidirectional images}\footnote{In this paper, we use 360° images or 360° data interchangeably.} or \textit{panoramic image} has emerged and attracted increasing attention from academia. Essentially, 360° images are spherical data captured by a camera from all angles and then projected onto a 2D image plane through equirectangular projection, resulting in ERP images (as shown in Fig.~\ref{fig:motivation}). Detecting salient objects in 360° data, known as \textit{360 SOD}, is arguably one of the most practical computer vision tasks in the 360° field and holds great significance in 360 recognition, 360 object detection and other related applications. Unfortunately, compared to 2D SOD, 360 SOD has received less attention, and there is still much to be explored in this field.

Given the similarity between 2D SOD and 360 SOD, one possible approach is to directly apply 2D SOD methods to 360 SOD. Currently, 2D SOD methods can be categorized into two types based on their foundation models: convolution-based SOD methods and Transformer-based SOD methods. Convolution-based SOD methods, such as EGNet~\cite{EGNet} and GateNet~\cite{GateNet}, typically utilize an encoder-decoder structure and have achieved impressive results. With the emergence of Transformer~\cite{transformer}, Transformer-based SOD methods, such as VST~\cite{VST}, have been proposed and achieved comparable results. These 2D SOD methods can deliver reasonable performance when applied to 360 SOD.

However, applying the aforementioned 2D SOD methods to 360 SOD without considering the unique characteristics of 360° data cannot further improve performance. 
As depicted in Fig.~\ref{fig:motivation}, due to the intrinsic spherical nature of 360° data, ERP images will exhibit distortions after undergoing the equirectangular projection process. Specifically, there are two main types of distortions: First, small regions near the poles on the original sphere will be stretched on an ERP image; Second, 
large objects that appear near the equator will exhibit a fisheye effect on an ERP image,  such as the castle in the example of Fig.~\ref{fig:motivation}.

Existing methods that are specifically designed for 360 SOD focus on handling the first type of distortion while ignoring the second one. One solution is to integrate a distortion adaptation module into the 2D SOD method, \textit{e.g.,} DDS~\cite{360-SOD}. They divide and merge the ERP image by image cropping or segmentation to reduce the effect of distortions. However, the improvement achieved by these methods is very limited. Another solution is to design a new projection strategy that avoids distortions. Specifically, FANet~\cite{FANet} and MPFR-Net~\cite{cong2023multi} use a cube-projection-based method, which results in smaller geometric deformations. Unfortunately, the cube-projection-based method may cause severe object-splitting problems that affect the object's integrity due to boundary discontinuity issues~\cite{cheng2018cube, Wang_2020_CVPR}. Additionally, we observed that the cube-projection-based method \cite{FANet} may suffer from poor training stability. 

To address the aforementioned problems, we propose a novel distortion-aware Transformer-based model for 360 SOD, named DATFormer. The model is essentially an encoder-decoder structure. Following the recent method proposed for 2D SOD called Visual Saliency Transformer (VST)~\cite{VST}, we adopt the Transformer model named T2T ViT~\cite{yuan2021tokens}, which has been pre-trained on ImageNet, as the encoder and the Reverse T2T of VST as the decoder. This design is intended to leverage the knowledge of the pre-trained model learned from 2D natural images. Our proposed model has two key components that account for the distortions in ERP images introduced by equirectangular projection: First, we design two distortion-adaptive modules in the decoder: the Distortion Mapping Module and the Distortion-adaptive Attention Block. Second, we introduce a novel relation matrix that provides prior information to the model. Specifically, we incorporate the Distortion Mapping Module equipped with deformable convolution between the encoder and decoder. This module guides the model to globally pre-adapt to distorted features extracted by the encoder. 
Building on this pre-adaptation, the Distortion-adaptive Attention Block in the decoder further alleviates local distortions on multi-scale features to enable the model to generate accurate result. Furthermore, we observed that the fisheye effect is more severe for large objects that are closer to the equator area in the original 360° images. Therefore, we encode this prior information as a relation matrix and feed it into the model along with positional embedding. We show that the relation matrix is beneficial for detecting objects, especially for those affected by the second type of distortion mentioned above. We conduct extensive experiments on three public 360 SOD datasets, including the 360-SOD dataset~\cite{360-SOD}, the F-360iSOD-A dataset~\cite{F-360iSOD}, and the 360-SSOD dataset~\cite{360-SSOD}. The experimental results demonstrate the effectiveness of these new modules and the superiority of our proposed model over existing state-of-the-art methods. 

In summary, the contributions of this work are three-fold: 

\begin{itemize}
    \item We propose a novel distortion-aware Transformer-based model, namely DATFormer, for 360 SOD.
    \item We introduce a Distortion Mapping Module and a Distortion-adaptive Attention Block that are specifically designed to handle distortions. Additionally, a novel relation matrix is proposed to represent the relationship between 360° data and ERP images. This relation matrix is incorporated as part of the positional embedding to further improve model performance.
    \item We evaluate the new model on three public 360 SOD datasets and compare it with existing 2D SOD and 360 SOD methods. The experimental results show that our new model outperforms the existing methods, achieving state-of-the-art performance on the 360 SOD task.
\end{itemize}
\section{Related Work}
\subsection{2D SOD Methods}
Over the past decade, deep learning approaches have dominated the state-of-the-art performance of 2D SOD, and CNN-based methods have become mainstream in SOD. 2D SOD methods usually adopt a framework of down-sampling or down-sampling combined with up-sampling to utilize the multi-scale feature. Among all CNN-based methods, most have attempted to develop new modules or structures. For example, \cite{li2017multi} proposed a multi-scale cascade structure that better consolidates contextual information and intermediate saliency priors. \cite{wang2017stagewise} proposed a framework with a stage-wise refinement mechanism and designed a pyramid pooling module to generate multi-level features. \cite{hou2017deeply} proposed a shortcut mechanism to integrate high-level features into low-level features and calculate the loss function with all the results of each scale. \cite{GateNet} proposed a simple gated network to adaptively control the amount of information between the encoder and decoder. Others have tried to find new forms of supervision. For example, both \cite{EGNet} and \cite{su2019selectivity} proposed to leverage edge map as guidance to improve the model's performance on the edges of salient objects. \cite{wu2019cascaded} designed a Partial Decoder and a Holistic Attention Module to generate an initial salient map as new supervision. \cite{zhao2020depth} proposed a new perspective of incorporating depth constraints into the learning process

In recent years, the Transformer model has gained popularity, and many researchers have attempted to apply it to visual tasks. 
Recent works such as
\cite{Vit, yuan2021tokens, liu2021swin} proposed several Transformer-based models and pre-trained on ImageNet~\cite{deng2009imagenet}. Among them, 
T2T ViT~\cite{yuan2021tokens} proposed a Tokens-to-Token module, which is a down-sampling process in essence. Based on this, a new Transformer-based SOD method called Visual Saliency Transformer(VST)~\cite{VST} was developed. VST used the pre-trained T2T ViT model as an encoder and designed a reverse Tokens-to-Token module, which is an up-sampling process in the decoder. Besides, VST leveraged both the salient map and edge map as supervision and designed a salient token and contour token to generate each map as a query. VST not only uses a relatively simple structure, but also achieves remarkable performance. 

\subsection{Methods on 360° Image}
Distortion is a common problem in the 360° field, and there are two main categories of methods proposed to address it. The first category is to directly design a distortion adaptation module to accommodate the ERP images. For example, \cite{zhang2022bending} proposed a Deformable Patch Embedding(DPE) module and a Deformable MLP(DMLP) module to make their model more adaptable to irregular deformation in ERP images. \cite{tateno2018distortion} proposed distortion-aware convolution filters for dense prediction in ERP images. \cite{guerrero2020s} replaced standard convolution with equirectangular convolutions, which are specifically designed for projected ERP images. The second category of methods to address distortion in 360° images is designing new projection methods. There are mainly four different types of projection: equirectangular projection, cube projection~\cite{cheng2018cube}, icosahedron projection~\cite{zhang2019orientation, lee2019spherephd, yoon2022spheresr}, and tangent projection~\cite{li2022omnifusion}. Currently, equirectangular projection is still the most commonly used approach.

Though many methods are proposed to address distortion issue in 360° images, few are designed for 360 SOD. \cite{360-SOD} proposed a distortion adaptation module, which divides the original image into blocks and then integrates them. \cite{FANet} proposed a model that separately extracts features from equirectangular projection and cube projection and then fuses them. \cite{cong2023multi} proposed a better cube-projection-based method than \cite{FANet}. Moreover, \cite{360-SSOD, MFFPANet} proposed multi-stage models to tackle 360 SOD. Furthermore, \cite{ODI-SOD} proposed a Transformer-based method while \cite{huang2023lightweight} proposed a lightweight method. Although many of them  achieve good results, some are either too cumbersome \cite{360-SSOD, FANet} or ineffective for the distortion problem like \cite{360-SOD, ODI-SOD}. It is worth noting that 360 SOD is different from the task of 360 object detection (OD). Specifically, 360 OD is class-specific while 360 SOD is class-agnostic, focusing on separating foreground objects from the background without recognizing their precise categories. 360 SOD is more sensitive to distortions than 360 OD.

Several methods attempt to incorporate spatial priors in ERP images for 360 tasks. \cite{liao2020model} addressed the Distortion Rectification task by training a distortion distribution map. \cite{nishiyama2021360} proposed a distortion map and concatenated it with the original input. \cite{shen2022panoformer} tackled the distortion problem by establishing a one-to-one correspondence between the ERP image and the spherical position. In this work, we propose a new strategy to exploit the spatial prior for 360 SOD.

\section{Methodology}
\begin{figure*}[th]
\centering
\includegraphics[width=0.95\linewidth]{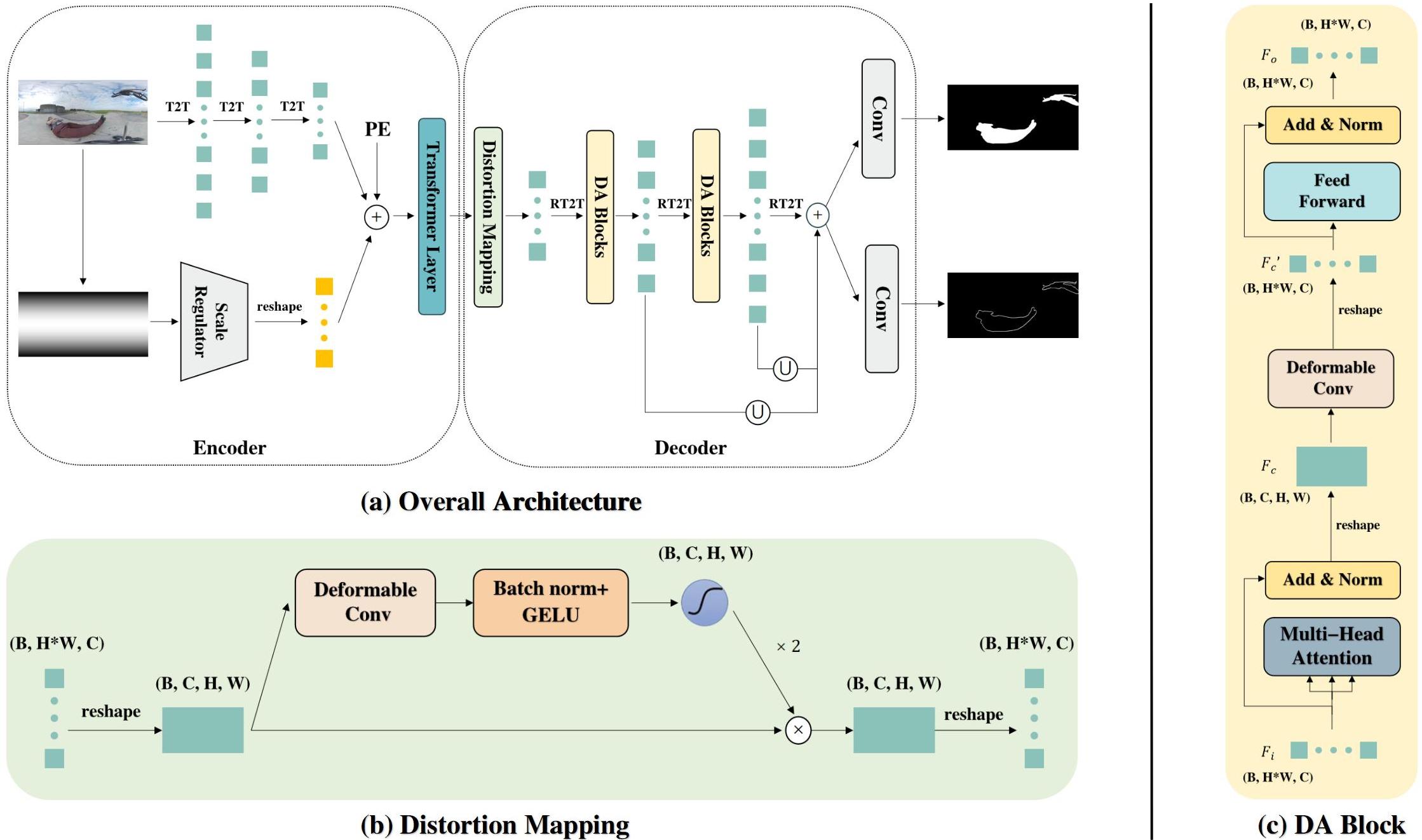}
\caption{(a) Overview of our framework. The ERP image is down-sampled through three T2T operations. Meanwhile, the relation matrix, which has the same size as the input 360° image, goes through a scale regulator module for dimension matching. The final features are generated by adding the down-sampled features and relation matrix with positional embedding and then passing through the Transformer layers. Before upsampling, the features pass through (b) a Distortion Mapping Module (DM) to adapt to the distortion. Upsampling is then performed through RT2T with (c) the Distortion-adaptive Attention Block (DA). Finally, the saliency map and edge map are predicted by the saliency detection head and edge detection head, respectively.}
\label{fig:pipeline}
\end{figure*}

\subsection{Overall Architecture}
\label{subsec:overview}
The goal of this work is to perform salient object detection (SOD) on ERP images converted from 360° data. The input is an ERP image and the output is its corresponding salient map. The overall architecture of our proposed DATFormer is illustrated in Fig.~\ref{fig:pipeline}. Similar to VST~\cite{VST}, the model comprises two parts: the encoder and the decoder, which correspond to the down-sampling and the up-sampling processes, respectively. Given that the aspect ratio of an ERP image is usually 1:2, which is different from ordinary images due to the 180°:360° ratio of the latitude and longitude in a sphere, we first adapt the structure of T2T ViT as our encoder and then use the Reverse T2T as our decoder. 
To address the distortion problem, we introduce two distortion adaptation modules in the decoder: a Distortion Mapping (DM) module and a Distortion-adaptive Attention (DA) block. These modules are designed to alleviate the distortions that occur in ERP images. Furthermore, to account for the location distribution of objects in 360° data, we present a Relation Matrix (RM) to encode this prior information and feed it into the model along with positional embedding.


\subsection{Modified T2T and Reverse T2T}
\label{subsec:t2t}
The encoder of our proposed DATFormer model consists of three T2T blocks and a Transformer layer with 14 Transformer blocks~\cite{transformer}. Given an ERP image with dimensions of $H \times W$, the encoder downsamples it to dimensions of $\frac{H}{16} \times \frac{W}{16}$. 
T2T is a Tokens-to-Token module proposed by T2T ViT~\cite{yuan2021tokens}. The T2T module employs a sliding window to split overlapped patches, referred to as \textit{soft split}. Two self-attention layers are positioned before and after the second soft split. Specifically, among the three soft split steps, the sliding window sizes are set to be \(k = [7, 3, 3]\), with stride of \(s = [3, 1, 1]\), and padding size of \(p = [2, 1, 1]\), respectively. Therefore, we can acquire multi-level tokens \(T_1\in\mathbb{R}^{l_1 \times c}\), \(T_2\in\mathbb{R}^{l_2 \times c}\) and \(T_3\in\mathbb{R}^{l_3 \times c}\), where \(l_1=\frac{H}{4}\times\frac{W}{4}\), \(l_2=\frac{H}{8} \times \frac{W}{8}\) and \(l_3=\frac{H}{16} \times \frac{W}{16}\), and $H$ and $W$ represent the height and width of the input ERP images. Follow the appraoch VST~\cite{VST}, we set \(c = 64\) and use a linear projection layer on \(T_3\) to transform its embedding dimension from \(c\) to \(d = 384\). 
In VST, both $H$ and $W$ are set to be 224. However, given that the aspect ratio of an ERP image is typically 1:2, we set the height of the input image to 224 and the width to 448. The encoder is initialized using T2T ViT that has been pre-trained on the ImageNet dataset.

The decoder of our model consists of a DM module, three Reverse T2T (RT2T) blocks, two DA modules, and two output heads, one for saliency detection  and the other for edge detection. The decoder progressively restores the feature map from $\frac{H}{16} \times \frac{W}{16}$ back to $H \times W$. RT2T is the inverse of T2T and upsamples tokens by folding each high-level token into a low-level token. The three RT2T modules correspond to three T2T modules with sliding window sizes of \(k = [3, 3, 7]\), \(s = [1, 1, 3]\), and \(p = [1, 1, 2]\). To ensure channel alignment, all embedding dimensions are projected from \(c = 64\) to \(d = 384\). 

\subsection{Distortion Mapping}
\label{subsec:DM}
Inspired by \cite{woo2018cbam}, we propose a Distortion Mapping (DM) Module as our first distortion-adaptive module. As shown in Fig.~\ref{fig:pipeline}(b), the DM module divides the feature map extracted by the encoder into tokens. To make the model more globally adaptable to the distortion of ERP images, we utilize deformable convolution~\cite{DCNN} to learn a distortion-adaptive matrix that represents the importance of each position. We then multiply this matrix with the original features to emphasize important features. The formula for the DM is as follows:
\begin{equation}
    DM = 2 \times \sigma(GELU(BN(DConv(F_i))))
\end{equation}
\begin{equation}
    F_o = DM \otimes F_i
\end{equation}
where \(F_i\) represents the input features extracted by the encoder, \(F_o\) indicates the output of this module, and DConv represents deformable convolution. Specifically, we adopt deformable convolution to extract the distortion features. Then we normalize this matrix to the range [0,2]. Values greater than 1 indicate more importance, while less than 1 is less important. Finally, we multiply this weight matrix with the original feature map to guide the model to pay attention to the more important part of the features.

We position the DM module right after the encoder to globally highlight the distorted regions in the features. Thus, the decoder has a pre-adaptation for distortion before the decoding process.

\begin{figure*}[t]
\centering
\includegraphics[width=\linewidth]{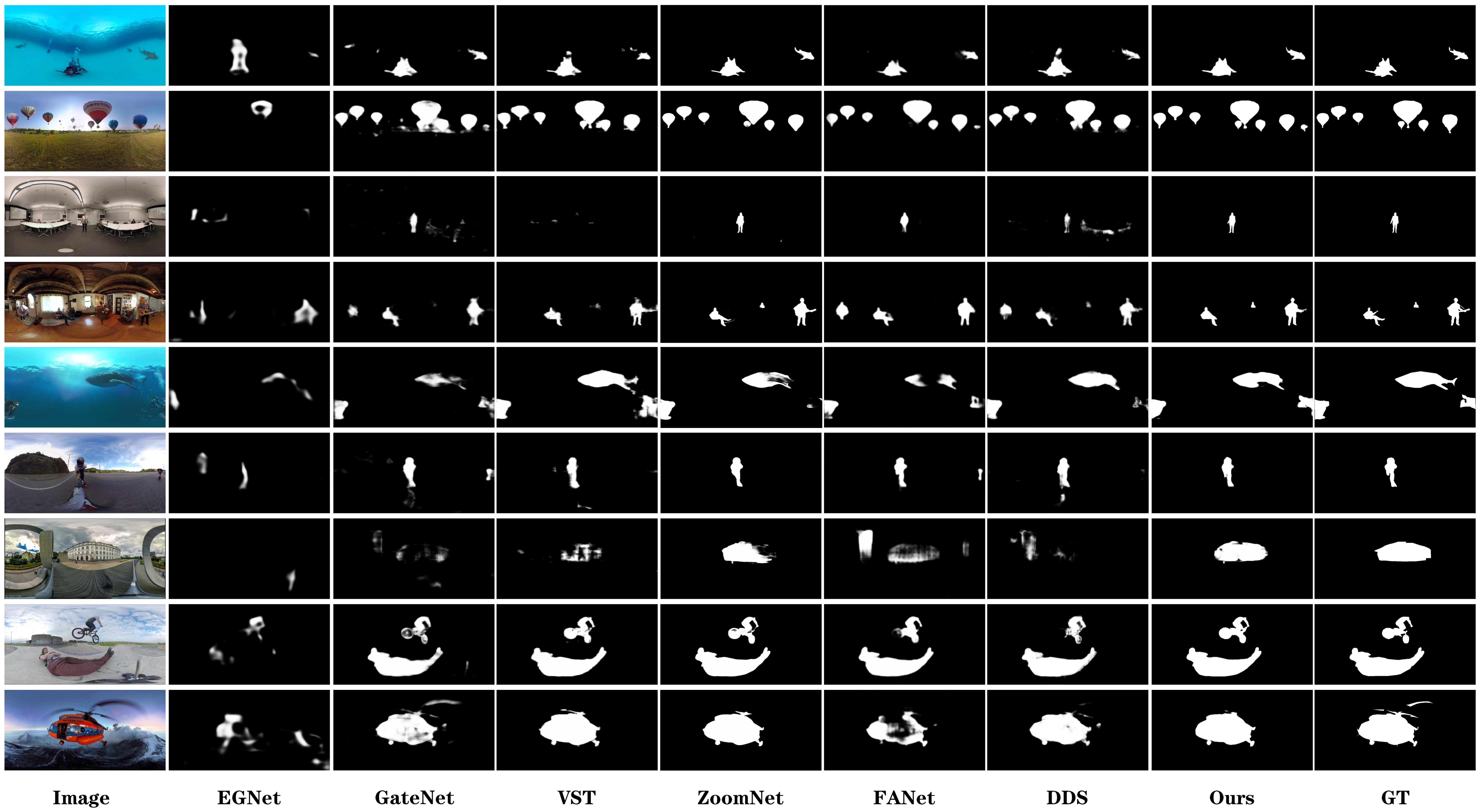}
\caption{Comparison of the qualitative results achieved by different methods on the 360-SOD dataset.}
\label{fig:vis}
\end{figure*}

\subsection{Distortion-adaptive Attention Block}
\label{subsec:DA}
Apart from using the DM module for pre-adaptation, we also attempted to adapt the Transformer block~\cite{transformer} to distortions without significantly modifying its original structure. As shown in Fig.~\ref{fig:pipeline}, we propose a Distortion-adaptive Attention (DA) Block by inserting a deformable convolution before the feedforward layers. Although the image is divided into tokens using sliding windows, it still retains its relative positional characteristics. Therefore, it is easy to restore the tokens back to the image size and incorporate convolution-based modules into the Transformer block. 
The formula is as follows:
\begin{equation}
    F_c = Reshape(MSA(F_i))
\end{equation}
\begin{equation}
    F_c' = Reshape(DConv(F_c)) 
\end{equation}
\begin{equation}
    F_o = MLP(F_c')
\end{equation}
where ``MSA" stands for Multi-head Self-Attention, ``Reshape" refers to a reshaping operation and ``DConv" indicates deformable convolution. As shown in Fig. ~\ref{fig:pipeline}(c), we insert the deformable convolution in between the MSA and MLP. Specifically, the input token \(F_i\) of MSA has dimensions of \([B, H*W, C]\) (where $B$ is the batch size, $H$ and $W$ represent the height and width of the features, respectively. $C$ denotes the number of channels). The input feature map \(F_c\) of DConv has dimensions of \([B, C, H, W]\). The input \(F_c'\) and output \(F_o\) of MLP have the same dimensions of \([B, H*W, C]\). To ensure that each module has a compatible size, we reshape the one-dimensional tokens into two-dimensional features after the MSA and then revert the two-dimensional features back to one-dimensional tokens after the DConv. Therefore, we can insert DConv in the middle of the attention module to consider the distortion information. 

During the up-sampling, we replace the regular attention blocks with the two DA Blocks, which are inserted right after the first and second RT2T modules. This allows our model to effectively alleviate local distortions over different scales of features.

\begin{table*}[t]
\centering
\large
\begin{tabular}{cc|cccc|cc|c}
\hline
 & \multicolumn{1}{l|}{} & \multicolumn{4}{c|}{2D SOD Methods} & \multicolumn{3}{c}{360 SOD Methods} \\ 
 \cline{3-9} 
Datasets & Metric & EGNet\cite{EGNet} & GateNet\cite{GateNet} & VST\cite{VST} & ZoomNet\cite{Pang_2022_CVPR} & DDS\cite{360-SOD} & FANet\cite{FANet} &\ Ours \\
\hline
{360-SOD} & $MAE\downarrow$ & 0.0597 & 0.0226  & 0.0217 & 0.0223 & 0.0230 & \underline{0.0207} & \textbf{0.0174} \\
& $maxF\uparrow$ & 0.3194 & 0.7398 & 0.7193 & 0.7230 & 0.7241 & \underline{0.7699} & \textbf{0.7928}\\
& $meanF\uparrow$ & 0.2889 & 0.7119 & 0.6965 & 0.7103 & 0.6960 & \underline{0.7484} & \textbf{0.7742} \\
& $maxE\uparrow$ & 0.7324 & 0.8805 & 0.8794 & 0.8485 & \underline{0.9058} & 0.9001 & \textbf{0.9191}\\\
& $meanE\uparrow$ & 0.6550 & 0.8456 & 0.8607 & 0.8357 & 0.8653 & \underline{0.8727} & \textbf{0.9071}\\
& $S_m\uparrow$ & 0.5901 & 0.8053 & 0.8093 & 0.8050 & 0.8030 & \underline{0.8261} & \textbf{0.8493}\\
\hline

{F-360iSOD-A} & $MAE\downarrow$ & 0.0772 & 0.0371 & 0.0424 & \textbf{0.0359} & 0.0562 & 0.0388 & \underline{0.0361}\\
& $maxF\uparrow$ & 0.2010 & 0.5025 & 0.4927 & 0.5208 & 0.2578 & \underline{0.5414} & \textbf{0.5455} \\
& $meanF\uparrow$ & 0.1739 & 0.4873  & 0.4707 & 0.5127 & 0.1556 & \underline{0.5242} & \textbf{0.5371}\\
& $maxE\uparrow$ & 0.6755 & 0.7596 & \textbf{0.8043} & 0.7558 & 0.6428 & \underline{0.7856} & 0.7661 \\
& $meanE\uparrow$ & 0.6003 & 0.6838 & 0.7238 & 0.6859 & 0.3867 & \textbf{0.7355} & \underline{0.7295} \\
& $S_m\uparrow$  & 0.6032 & 0.7628 & 0.7403 & 0.7526 & 0.5779 & \textbf{0.7901} & \underline{0.7664}\\
\hline

{360-SSOD} & $MAE\downarrow$ & 0.0618 & 0.0280 & 0.0300 & \underline{0.0279} & 0.1862 & 0.0554 & \textbf{0.0261}\\
& $maxF\uparrow$ & 0.2605 & \underline{0.6381} & 0.6351 & 0.6198 & 0.1544 & 0.5525 & \textbf{0.6573} \\
& $meanF\uparrow$ & 0.2247 & \underline{0.6100} & 0.5802 & 0.5946 & 0.1518 & 0.4929 & \textbf{0.6440} \\
& $maxE\uparrow$ & 0.7184 & \underline{0.8602} & 0.8565 & 0.8349 & 0.5803 & 0.7801 & \textbf{0.8648} \\
& $meanE\uparrow$ & 0.6181 & \underline{0.8143} & 0.7922 & 0.7869 & 0.4904 & 0.7241 & \textbf{0.8322} \\
& $S_m\uparrow$  & 0.5580 & \underline{0.7644} & 0.7549 & 0.7411 & 0.4623 & 0.6568 & \textbf{0.7698} \\
\hline
\end{tabular}
\caption{Comparison of the quantitative results achieved by our proposed model and six state-of-the-art methods on 360 SOD dataset, F-360iSOD dataset, and 360-SSOD dataset. All results were obtained through our experiments.}
\label{tab:results}
\end{table*}

\subsection{PE with Relation Matrix}
\label{subsec:RM}
Upon observing the 360° images, we noticed that the fisheye effect is more severe for large objects closer to the equator area in the original 360° image. As a result, these distorted objects mainly appear in the middle regions of the ERP images. To address this issue, we aim to incorporate this spatial prior into our model. 

We represent this spatial prior by taking the cosine of the latitude in 360° spherical data and projecting it into an ERP image to create an initialized relation matrix (RM). The cosine function's value is close to 1 when the latitude is closer to the equator and close to 0 otherwise, which allows the model to pay more attention to objects near the equator area. The relation matrix is initialized as Eq.~\ref{eq:RM}.
\begin{equation}
    matrix[\phi, \theta] = \cos \theta
    \label{eq:RM}
\end{equation}
where \(\phi \in (-\pi, \pi)\) and \(\theta \in (-\frac{\pi}{2}, \frac{\pi}{2})\) correspond to longitude and latitude on the sphere, respectively. The relationship between \(\phi, \theta\) and the coordinates of the ERP image is shown below.
\begin{equation}
    \phi =  \frac{w - W/2}{W} \times 2\pi,   \theta = \frac{h - H/2}{H} \times \pi
    \label{eq:RM_}
\end{equation}
where $H$, $W$ represent the height and width of the ERP image, respectively. $h$, and $w$ represent the vertical and horizontal coordinates of the ERP image, respectively.

In T2T ViT, the positional embeddings are added after the three down-sampling T2T modules. As shown in Fig.~\ref{fig:pipeline}, we design a simple CNN structure called Scale Regulator that performs down-sampling three times. Thus, the scale regulator can modify the size of the matrix to match the size of positional embedding. In addition, the scale regulator makes the relation matrix learnable instead of fixed. Finally, we add the relation matrix and positional embedding and then feed them into subsequent model layers.

\subsection{Objectives}
We use the binary cross-entropy loss as the loss function when training the saliency detection head and the edge detection head, which is calculated as Eq.~\ref{eq:loss}.
\begin{equation}
    L = - \frac{1}{W \times H} \sum_{w=1}^{W}\sum_{h=1}^{H}[G_w{}_h \ln P_w{}_h + (1-G_w{}_h)\ln(1-P_w{}_h)]
    \label{eq:loss}
\end{equation}
\noindent where $W$ and $H$ denote the width and height of the ERP image, respectively. For the saliency loss, \(G_w{}_h\) refers to the ground truth saliency score at location $(w, h)$, and $P_w{}_h$ denotes the predicted saliency score at location $(w, h)$. For the edge loss,  \(G_w{}_h\) and $P_w{}_h$ refer to ground truth and predicted edge intensity values at location $(w, h)$, respectively.

In addition to the saliency map, we also use the edge map as a supervision, similar to \cite{EGNet}. Edge maps are generated from saliency maps and only capture the edges of salient objects. Thus, the total loss function \(L_{total}\) is defined as Eq.~\ref{eq:total}.
\begin{equation}
    L_{total} = L_{sal} + L_{edge}
    \label{eq:total}
\end{equation}
\(L_{sal}\) and \(L_{edge}\) denote saliency map loss and edge loss, respectively.
\section{Experiment}

\subsection{Datasets and Evaluation Metrics}
\textbf{Datasets.}\quad
We evaluated our model on three publicly available datasets.
The first dataset, called the \textbf{360-SOD dataset}~\cite{360-SOD}, consists of 500 ERP images that are divided into 400 training images and 100 testing images. This dataset was collected from various omnidirectional video datasets, such as \cite{cheng2018cube,Salient!360}.
The second dataset, called \textbf{F-360iSOD dataset}~\cite{F-360iSOD}, comprises one training set and two testing sets: F-360iSOD-train, F-360iSODtestA, and F-360iSOD-testB. F-360iSODtrain and F-360iSODtestA contain 68 and 17 equirectangular images, respectively, from the Salient!360~\cite{Salient!360}, while F-360iSOD-testB contains 22 ERP images from another panoramic image dataset named Stanford360~\cite{Stanford360}.
The third dataset, named \textbf{360-SSOD dataset}~\cite{360-SSOD}, consists of 1105 equirectangular 360° images, including 850 training images and 255 testing images.

\begin{figure*}[t]
\centering
\includegraphics[width=0.95\linewidth]{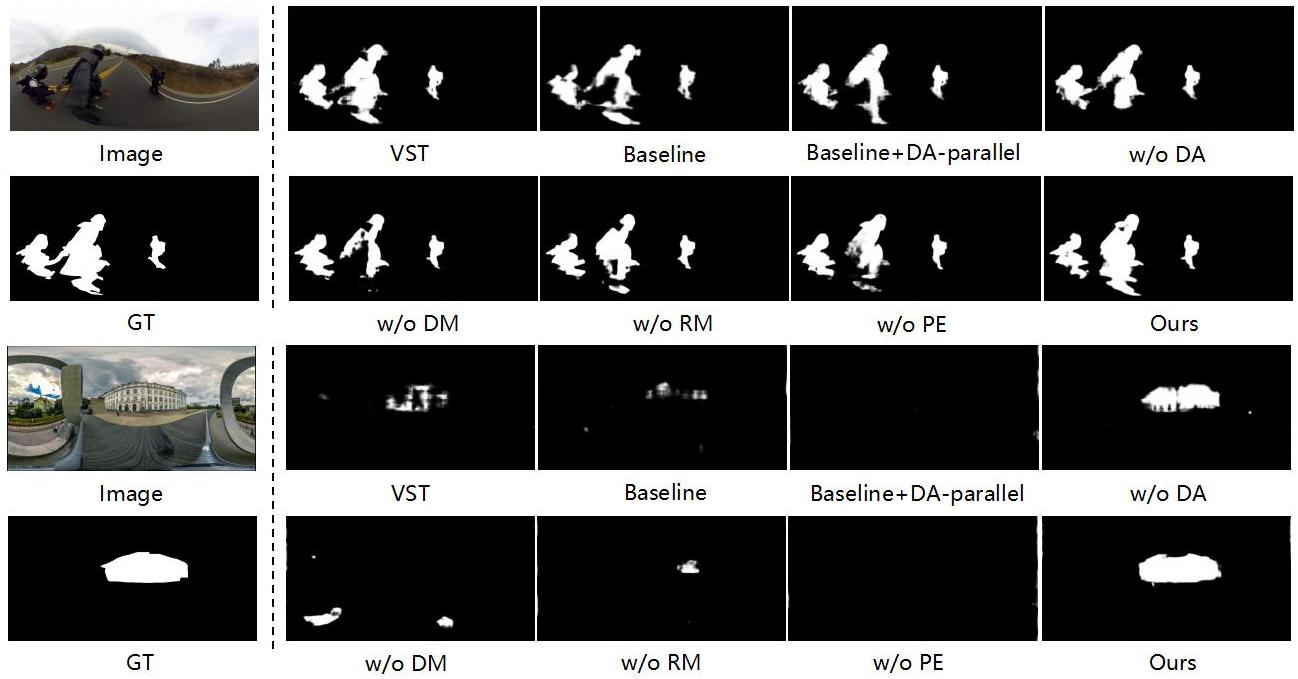}
\caption{Comparison of the qualitative results achieved by different model variants on the 360-SOD dataset. }
\label{fig:ablation}
\end{figure*}

\begin{table*}[t]
\centering
\large
\begin{tabular}{c|cccccccccc}
\hline
Metrics &  VST & Baseline & Baseline+DA-parallel & w/o DA & w/o DM & w/o PE & w/o RM & Ours \\
\hline
$MAE\downarrow$ & 0.0217 & 0.0203 & 0.0198 & 0.0193 & 0.0194 & 0.0199 & \underline{0.0184} & \textbf{0.0174} \\
\hline
$maxF\uparrow$ & 0.7193 & 0.7582 & 0.7676 & 0.7667 & 0.7766 & 0.7481 & \underline{0.7855} & \textbf{0.7928} \\
\hline
$meanF\uparrow$ & 0.6965 & 0.7375 & 0.7362 & 0.7487 & 0.7586 & 0.7306 & \underline{0.7724} & \textbf{0.7742} \\
\hline
$maxE\uparrow$ & 0.8794 & 0.8882 & 0.8972 & \underline{0.9062} & 0.8932 & 0.8774 & 0.9005 & \textbf{0.9191} \\
\hline
$meanE\uparrow$ & 0.8607 & 0.8681 & 0.8799 & \underline{0.8908} & 0.8829 & 0.8617 & 0.8871 & \textbf{0.9071} \\
\hline
$S_m\uparrow$ & 0.8093 & 0.8189 & 0.8341 & 0.8399 & \underline{0.8459} & 0.8310 & 0.8436 & \textbf{0.8493} \\
\hline
\end{tabular}
\caption{Ablation analysis of the proposed model on the 360-SOD dataset. }
\label{tab:ablation}
\end{table*}

\noindent\textbf{Evaluation Metrics.}\quad
To evaluate our proposed model, we used four quantitative evaluation metrics to compare our method with other baseline methods, including Structure measure \cite{S-measure}, F-measure, E-measure \cite{E-measure}, and Mean Absolute Error (MAE)~\cite{MAE}. Specifically, Structure measure ($S_m$) evaluates the region-aware and object-aware structural similarity of the predicted saliency map with the ground truth. F-measure considers both precision and recall. Enh-anced-alignment measure (E-measure) takes into account pixel-level and image-level errors simultaneously.  Mean Absolute Error (MAE) computes pixel-wise average absolute error. For F-measure and E-measure, we also calculated the maximum value and average value of the metric across all test images.

\subsection{Implementation Details}
Considering the domain bias of the three datasets, we trained and tested the models separately on the three datasets. We followed the way used in \cite{EGNet} to generate contour maps from the GT saliency map to compute edge loss. To augment the training data, we used random cropping and random flipping. Specifically, we first resized each ERP image to \(512 \times 256\) and then randomly cropped image regions of size \(448 \times 224\) as the model input. Finally, we randomly flipped the image to further increase the diversity of the training data.

During the training process, we modified the pre-trained \(T2T-ViT_t-14\) model by doubling the parameters of positional embedding to match the input size of \(448 \times 224\). We set the batch size to 4 and the total training steps to 70,000. To optimize the model, we adopted the Adam optimizer~\cite{Adam} . The initial learning rate is set to 0.0001 and is decayed to the previous value of 0.1 at 45,000 steps and 60,000 steps. We implemented our model using Pytorch \cite{Pytorch} and trained it on an RTX 2080Ti GPU.

\subsection{Comparison with Baseline Methods}
We compared our model with six state-of-the-art CNN-based SOD models, including four 2D SOD models: EGNet\cite{EGNet}, GateNet\cite{GateNet}, VST\cite{VST}, ZoomNet\cite{Pang_2022_CVPR}, as well as two 360 SOD models: DDS\cite{360-SOD} and FANet\cite{FANet}. For 2D SOD methods and FANet, we used the respective models' default parameters for training and then fine-tuned them to find the best experimental results. For DDS, we re-implemented the code of DDS using PyTorch and re-trained the model on F-360iSOD and 360-SSOD datasets.

As shown in Table~\ref{tab:results}, our proposed model outperforms all previous state-of-the-art 2D and 360 SOD models on 360-SOD and 360-SSOD datasets. On the F-360iSOD dataset, our model achieves comparable results as other models. As illustrated in Fig.~\ref{fig:vis}, our model produces better detection results for objects with or without distortions. Specifically, our model shows better integrity in detecting large and distorted objects, and performs better in capturing the details of small objects. Furthermore, we observed that our model is capable of detecting objects that are separated on both sides of the image, while other models either lack integrity or details in such cases. This can be attributed to the fact that our model is equipped with distortion-adaptive modules and a relation matrix, which enable it to handle challenging cases more effectively. 

\begin{figure*}[t]
\centering
\includegraphics[width=0.95\linewidth]{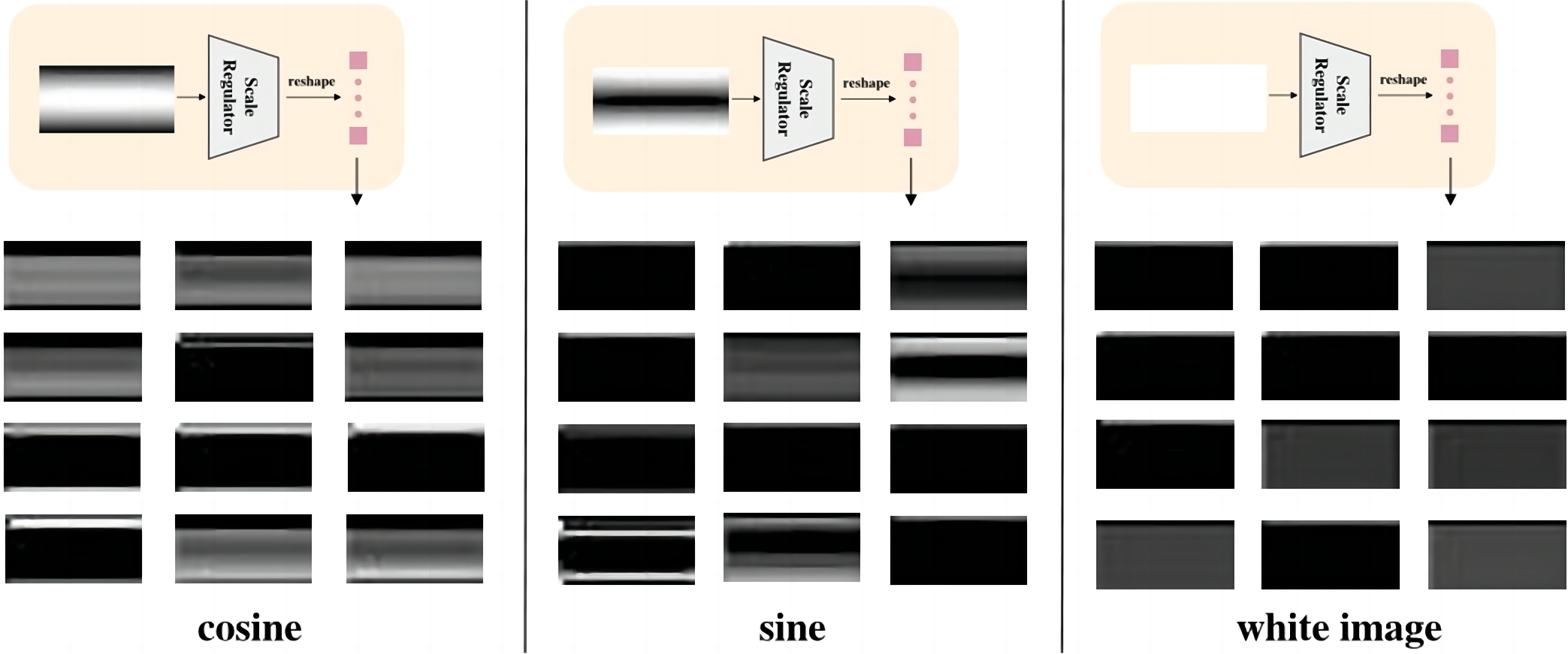}
\caption{Visualization of Relation Matrix with different prior maps after the Scale Regulator. The bottom figures show the outputs of the first 12 channels.}
\label{fig:RM_vis_v2}
\end{figure*}
\begin{table*}[t]
\centering
\large
\begin{tabular}{c|cccc}
\hline
Metrics & w/o RM & RM with Sine & RM with blank image & Ours \\
\hline
$MAE\downarrow$ & \underline{0.0184} & 0.0193 & 0.0194 & \textbf{0.0174} \\
\hline
$maxF\uparrow$ & \underline{0.7855} & 0.7653 & 0.7756 & \textbf{0.7928} \\
\hline
$meanF\uparrow$ & \underline{0.7724} & 0.7470 & 0.7577 & \textbf{0.7742} \\
\hline
$maxE\uparrow$ & \underline{0.9005} & 0.8853 & 0.8943 & \textbf{0.9191} \\
\hline
$meanE\uparrow$ & \underline{0.8871} & 0.8746 & 0.8829 & \textbf{0.9071} \\
\hline
$S_m\uparrow$ & \underline{0.8436} & 0.8375 & 0.8406 & \textbf{0.8493} \\
\hline
\end{tabular}
\caption{Comparison of results obtained on the 360-SOD dataset by using different prior maps in the relation matrix. }
\label{tab:RM_ablation}
\end{table*}

\subsection{Ablation Study}
In this section, we will demonstrate the effectiveness of the individual components. The experimental results on 360-SOD dataset are shown in Tab. ~\ref{tab:ablation}.

\noindent\textbf{Modified T2T and RT2T.}\quad
Although both our baseline model and VST use the T2T and RT2T modules, we modified and expanded the T2T module to adapt it to the input size of the ERP images (as described in Sec.~\ref{subsec:t2t}). As shown in Table~\ref{tab:ablation}, our baseline model has significantly improved the results compared to the VST model, demonstrating the effectiveness of our modified structure. However, while both VST and baseline methods can detect a few objects in the ERP images, they perform relatively poorly on distorted objects compared with other variants, as illustrated in Fig.~\ref{fig:ablation}.

\noindent\textbf{Distortion Adaptive Modules.}\quad
In our proposed model, we introduced two distortion adaptation modules, Distortion-adaptive Attention Block (DA) and Distortion Mapping (DM). As shown in Table~\ref{tab:ablation}, we evaluated the performance of different variants by removing these two modules individually ("w/o DA" and "w/o DM") and observed that each module contributes to the overall performance of the model. As illustrated in Fig.~\ref{fig:ablation}, the detection results of "w/o DA" and "w/o DM" are much worse than our final model, particularly for the cases with  fisheye effect distortions.

Additionally, we experimented with incorporating our DA module into the Encoder of the baseline model and making it parallel with the Transformer layer, denoted as "Baseline+DA-parallel". However, through our experiments, we found that while adding the DA module in the Encoder did have a certain effect compared to the Baseline model, its performance was not as good as that achieved by using the DA module in the Decoder (\textit{i.e.}, our final model).

\noindent\textbf{Relation Matrix.}\quad
In our proposed model, a relation matrix is introduced to encode the spatial prior in the ERP images. As shown in Tab. ~\ref{tab:ablation}, the comparison between "Ours" and "w/o RM" demonstrates that adding the relation matrix to positional embedding can further improve performance. Furthermore, by comparing the results achieved by the model variant "w/o PE" and "Ours", we can see that the relation matrix is a complement to the original positional embedding rather than a replacement. 

Besides, 
we compared the effect of using  different prior maps in RM. The reason we chose the cosine function is due to the spatial distribution of objects in 360° data. 
Our statistical analysis revealed that approximately 85\% of the objects in 360-SOD dataset are concentrated in the center, where these objects are distorted to varying degrees by the fisheye effect. The performance of the model variants of replacing cosine function with a sine function or a blank image are shown in Tab. ~\ref{tab:RM_ablation}. Both the sine function and blank image have a negative effect on the model, suggesting that these spatial priors are not suitable. Moreover, as shown in Fig.~\ref{fig:RM_vis_v2}, different initialized relation matrix represent different spatial priors, and the features learned by scale regulator strongly depend on the initialized relation matrix. Appropriate spatial priors, such as using cosine function, can guide the model to pay attention to the expected regions of interest.
\section{Conclusion}

In this paper, we have explored the characteristics of ERP images and developed a Transformer-based model for 360 SOD. To handle the distortion problem of ERP images, we proposed two distortion adaptive modules, named Distortion Mapping and Distortion-adaptive Attention block, to make the model more adaptable to the distortion in ERP images. We also designed a relation matrix to formulate spatial prior of 360° data and used it along with positional embedding. The experimental results show that our model outperforms existing 2D SOD and 360 SOD methods, achieving new state-of-the-art performance.

\noindent\textbf{Acknowledgement}\quad This work was supported in part by The
Hong Kong Jockey Club Charities Trust under Grant 2022-0174, in part by
the Startup Funding and the Seed Funding for Basic Research for New Staff
from The University of Hong Kong, in part by the funding from UBTECH
Robotics, and in part by the National
Natural Science Foundation of China (No.~62002012 and No.~62006012).

{
\bibliographystyle{ACM-Reference-Format}
\bibliography{bibfile}


\begin{thebibliography}{45}


\ifx \showCODEN    \undefined \def \showCODEN     #1{\unskip}     \fi
\ifx \showDOI      \undefined \def \showDOI       #1{#1}\fi
\ifx \showISBNx    \undefined \def \showISBNx     #1{\unskip}     \fi
\ifx \showISBNxiii \undefined \def \showISBNxiii  #1{\unskip}     \fi
\ifx \showISSN     \undefined \def \showISSN      #1{\unskip}     \fi
\ifx \showLCCN     \undefined \def \showLCCN      #1{\unskip}     \fi
\ifx \shownote     \undefined \def \shownote      #1{#1}          \fi
\ifx \showarticletitle \undefined \def \showarticletitle #1{#1}   \fi
\ifx \showURL      \undefined \def \showURL       {\relax}        \fi
\providecommand\bibfield[2]{#2}
\providecommand\bibinfo[2]{#2}
\providecommand\natexlab[1]{#1}
\providecommand\showeprint[2][]{arXiv:#2}

\bibitem[Borji et~al\mbox{.}(2015)]%
        {MAE}
\bibfield{author}{\bibinfo{person}{Ali Borji}, \bibinfo{person}{Ming-Ming
  Cheng}, \bibinfo{person}{Huaizu Jiang}, {and} \bibinfo{person}{Jia Li}.}
  \bibinfo{year}{2015}\natexlab{}.
\newblock \showarticletitle{Salient object detection: A benchmark}.
\newblock \bibinfo{journal}{\emph{IEEE transactions on image processing}}
  \bibinfo{volume}{24}, \bibinfo{number}{12} (\bibinfo{year}{2015}),
  \bibinfo{pages}{5706--5722}.
\newblock


\bibitem[Chen et~al\mbox{.}(2023)]%
        {MFFPANet}
\bibfield{author}{\bibinfo{person}{Gang Chen}, \bibinfo{person}{Feng Shao},
  \bibinfo{person}{Xiongli Chai}, \bibinfo{person}{Qiuping Jiang}, {and}
  \bibinfo{person}{Yo-Sung Ho}.} \bibinfo{year}{2023}\natexlab{}.
\newblock \showarticletitle{Multi-Stage Salient Object Detection in
  $360^{\circ}$ Omnidirectional Image Using Complementary Object-Level Semantic
  Information}.
\newblock \bibinfo{journal}{\emph{IEEE Transactions on Emerging Topics in
  Computational Intelligence}} (\bibinfo{year}{2023}), \bibinfo{pages}{1--14}.
\newblock


\bibitem[Cheng et~al\mbox{.}(2018)]%
        {cheng2018cube}
\bibfield{author}{\bibinfo{person}{Hsien-Tzu Cheng}, \bibinfo{person}{Chun-Hung
  Chao}, \bibinfo{person}{Jin-Dong Dong}, \bibinfo{person}{Hao-Kai Wen},
  \bibinfo{person}{Tyng-Luh Liu}, {and} \bibinfo{person}{Min Sun}.}
  \bibinfo{year}{2018}\natexlab{}.
\newblock \showarticletitle{Cube Padding for Weakly-Supervised Saliency
  Prediction in $360^{\circ}$ Videos}. In \bibinfo{booktitle}{\emph{Proceedings
  of the IEEE Conference on Computer Vision and Pattern Recognition (CVPR)}}.
\newblock


\bibitem[Cong et~al\mbox{.}(2023)]%
        {cong2023multi}
\bibfield{author}{\bibinfo{person}{Runmin Cong}, \bibinfo{person}{Ke Huang},
  \bibinfo{person}{Jianjun Lei}, \bibinfo{person}{Yao Zhao},
  \bibinfo{person}{Qingming Huang}, {and} \bibinfo{person}{Sam Kwong}.}
  \bibinfo{year}{2023}\natexlab{}.
\newblock \showarticletitle{Multi-Projection Fusion and Refinement Network for
  Salient Object Detection in $360^{\circ}$ Omnidirectional Image}.
\newblock \bibinfo{journal}{\emph{IEEE Transactions on Neural Networks and
  Learning Systems}} (\bibinfo{year}{2023}).
\newblock


\bibitem[Dai et~al\mbox{.}(2017)]%
        {DCNN}
\bibfield{author}{\bibinfo{person}{Jifeng Dai}, \bibinfo{person}{Haozhi Qi},
  \bibinfo{person}{Yuwen Xiong}, \bibinfo{person}{Yi Li},
  \bibinfo{person}{Guodong Zhang}, \bibinfo{person}{Han Hu}, {and}
  \bibinfo{person}{Yichen Wei}.} \bibinfo{year}{2017}\natexlab{}.
\newblock \showarticletitle{Deformable convolutional networks}. In
  \bibinfo{booktitle}{\emph{Proceedings of the IEEE International Conference on
  Computer Vision(ICCV)}}.
\newblock


\bibitem[Deng et~al\mbox{.}(2009)]%
        {deng2009imagenet}
\bibfield{author}{\bibinfo{person}{Jia Deng}, \bibinfo{person}{Wei Dong},
  \bibinfo{person}{Richard Socher}, \bibinfo{person}{Li-Jia Li},
  \bibinfo{person}{Kai Li}, {and} \bibinfo{person}{Li Fei-Fei}.}
  \bibinfo{year}{2009}\natexlab{}.
\newblock \showarticletitle{Imagenet: A large-scale hierarchical image
  database}. In \bibinfo{booktitle}{\emph{Proceedings of the IEEE Conference on
  Computer Vision and Pattern Recognition(CVPR)}}.
\newblock


\bibitem[Dosovitskiy et~al\mbox{.}(2020)]%
        {Vit}
\bibfield{author}{\bibinfo{person}{Alexey Dosovitskiy}, \bibinfo{person}{Lucas
  Beyer}, \bibinfo{person}{Alexander Kolesnikov}, \bibinfo{person}{Dirk
  Weissenborn}, \bibinfo{person}{Xiaohua Zhai}, \bibinfo{person}{Thomas
  Unterthiner}, \bibinfo{person}{Mostafa Dehghani}, \bibinfo{person}{Matthias
  Minderer}, \bibinfo{person}{Georg Heigold}, \bibinfo{person}{Sylvain Gelly},
  {et~al\mbox{.}}} \bibinfo{year}{2020}\natexlab{}.
\newblock \showarticletitle{An image is worth 16x16 words: Transformers for
  image recognition at scale}.
\newblock \bibinfo{journal}{\emph{arXiv preprint arXiv:2010.11929}}
  (\bibinfo{year}{2020}).
\newblock


\bibitem[Fan et~al\mbox{.}(2017)]%
        {S-measure}
\bibfield{author}{\bibinfo{person}{Deng-Ping Fan}, \bibinfo{person}{Ming-Ming
  Cheng}, \bibinfo{person}{Yun Liu}, \bibinfo{person}{Tao Li}, {and}
  \bibinfo{person}{Ali Borji}.} \bibinfo{year}{2017}\natexlab{}.
\newblock \showarticletitle{Structure-measure: A new way to evaluate foreground
  maps}. In \bibinfo{booktitle}{\emph{Proceedings of the IEEE International
  Conference on Computer Vision(ICCV)}}.
\newblock


\bibitem[Fan et~al\mbox{.}(2018)]%
        {E-measure}
\bibfield{author}{\bibinfo{person}{Deng-Ping Fan}, \bibinfo{person}{Cheng
  Gong}, \bibinfo{person}{Yang Cao}, \bibinfo{person}{Bo Ren},
  \bibinfo{person}{Ming-Ming Cheng}, {and} \bibinfo{person}{Ali Borji}.}
  \bibinfo{year}{2018}\natexlab{}.
\newblock \showarticletitle{Enhanced-alignment measure for binary foreground
  map evaluation}.
\newblock \bibinfo{journal}{\emph{arXiv preprint arXiv:1805.10421}}
  (\bibinfo{year}{2018}).
\newblock


\bibitem[Guerrero-Viu et~al\mbox{.}(2020)]%
        {guerrero2020s}
\bibfield{author}{\bibinfo{person}{Julia Guerrero-Viu}, \bibinfo{person}{Clara
  Fernandez-Labrador}, \bibinfo{person}{C{\'e}dric Demonceaux}, {and}
  \bibinfo{person}{Jose~J Guerrero}.} \bibinfo{year}{2020}\natexlab{}.
\newblock \showarticletitle{What’s in my room? object recognition on indoor
  panoramic images}. In \bibinfo{booktitle}{\emph{Proceedings of the IEEE
  International Conference on Robotics and Automation (ICRA)}}.
\newblock


\bibitem[Hou et~al\mbox{.}(2017)]%
        {hou2017deeply}
\bibfield{author}{\bibinfo{person}{Qibin Hou}, \bibinfo{person}{Ming-Ming
  Cheng}, \bibinfo{person}{Xiaowei Hu}, \bibinfo{person}{Ali Borji},
  \bibinfo{person}{Zhuowen Tu}, {and} \bibinfo{person}{Philip~HS Torr}.}
  \bibinfo{year}{2017}\natexlab{}.
\newblock \showarticletitle{Deeply supervised salient object detection with
  short connections}. In \bibinfo{booktitle}{\emph{Proceedings of the IEEE
  Conference on Computer Vision and Pattern Recognition(CVPR)}}.
\newblock


\bibitem[Huang et~al\mbox{.}(2023)]%
        {huang2023lightweight}
\bibfield{author}{\bibinfo{person}{Mengke Huang}, \bibinfo{person}{Gongyang
  Li}, \bibinfo{person}{Zhi Liu}, {and} \bibinfo{person}{Linchao Zhu}.}
  \bibinfo{year}{2023}\natexlab{}.
\newblock \showarticletitle{Lightweight Distortion-aware Network for Salient
  Object Detection in Omnidirectional Images}.
\newblock \bibinfo{journal}{\emph{IEEE Transactions on Circuits and Systems for
  Video Technology}} (\bibinfo{year}{2023}).
\newblock


\bibitem[Huang et~al\mbox{.}(2020)]%
        {FANet}
\bibfield{author}{\bibinfo{person}{Mengke Huang}, \bibinfo{person}{Zhi Liu},
  \bibinfo{person}{Gongyang Li}, \bibinfo{person}{Xiaofei Zhou}, {and}
  \bibinfo{person}{Olivier Le~Meur}.} \bibinfo{year}{2020}\natexlab{}.
\newblock \showarticletitle{FANet: Features Adaptation Network for
  $360^{\circ}$ Omnidirectional Salient Object Detection}.
\newblock \bibinfo{journal}{\emph{IEEE Signal Processing Letters}}
  \bibinfo{volume}{27} (\bibinfo{year}{2020}), \bibinfo{pages}{1819--1823}.
\newblock


\bibitem[Kingma and Ba(2014)]%
        {Adam}
\bibfield{author}{\bibinfo{person}{Diederik~P Kingma} {and}
  \bibinfo{person}{Jimmy Ba}.} \bibinfo{year}{2014}\natexlab{}.
\newblock \showarticletitle{Adam: A method for stochastic optimization}.
\newblock \bibinfo{journal}{\emph{arXiv preprint arXiv:1412.6980}}
  (\bibinfo{year}{2014}).
\newblock


\bibitem[Lee et~al\mbox{.}(2019)]%
        {lee2019spherephd}
\bibfield{author}{\bibinfo{person}{Yeonkun Lee}, \bibinfo{person}{Jaeseok
  Jeong}, \bibinfo{person}{Jongseob Yun}, \bibinfo{person}{Wonjune Cho}, {and}
  \bibinfo{person}{Kuk-Jin Yoon}.} \bibinfo{year}{2019}\natexlab{}.
\newblock \showarticletitle{Spherephd: Applying cnns on a spherical polyhedron
  representation of 360deg images}. In \bibinfo{booktitle}{\emph{Proceedings of
  the IEEE/CVF Conference on Computer Vision and Pattern Recognition(CVPR)}}.
\newblock


\bibitem[Li et~al\mbox{.}(2020)]%
        {360-SOD}
\bibfield{author}{\bibinfo{person}{Jia Li}, \bibinfo{person}{Jinming Su},
  \bibinfo{person}{Changqun Xia}, {and} \bibinfo{person}{Yonghong Tian}.}
  \bibinfo{year}{2020}\natexlab{}.
\newblock \showarticletitle{Distortion-Adaptive Salient Object Detection in
  360$^\circ$ Omnidirectional Images}.
\newblock \bibinfo{journal}{\emph{IEEE Journal of Selected Topics in Signal
  Processing}} \bibinfo{volume}{14}, \bibinfo{number}{1}
  (\bibinfo{year}{2020}), \bibinfo{pages}{38--48}.
\newblock


\bibitem[Li et~al\mbox{.}(2017)]%
        {li2017multi}
\bibfield{author}{\bibinfo{person}{Xin Li}, \bibinfo{person}{Fan Yang},
  \bibinfo{person}{Hong Cheng}, \bibinfo{person}{Junyu Chen},
  \bibinfo{person}{Yuxiao Guo}, {and} \bibinfo{person}{Leiting Chen}.}
  \bibinfo{year}{2017}\natexlab{}.
\newblock \showarticletitle{Multi-scale cascade network for salient object
  detection}. In \bibinfo{booktitle}{\emph{Proceedings of the 25th ACM
  international conference on Multimedia}}.
\newblock


\bibitem[Li et~al\mbox{.}(2022)]%
        {li2022omnifusion}
\bibfield{author}{\bibinfo{person}{Yuyan Li}, \bibinfo{person}{Yuliang Guo},
  \bibinfo{person}{Zhixin Yan}, \bibinfo{person}{Xinyu Huang},
  \bibinfo{person}{Ye Duan}, {and} \bibinfo{person}{Liu Ren}.}
  \bibinfo{year}{2022}\natexlab{}.
\newblock \showarticletitle{Omnifusion: 360 monocular depth estimation via
  geometry-aware fusion}. In \bibinfo{booktitle}{\emph{Proceedings of the
  IEEE/CVF Conference on Computer Vision and Pattern Recognition(CVPR)}}.
\newblock


\bibitem[Liao et~al\mbox{.}(2020)]%
        {liao2020model}
\bibfield{author}{\bibinfo{person}{Kang Liao}, \bibinfo{person}{Chunyu Lin},
  \bibinfo{person}{Yao Zhao}, {and} \bibinfo{person}{Mai Xu}.}
  \bibinfo{year}{2020}\natexlab{}.
\newblock \showarticletitle{Model-free distortion rectification framework
  bridged by distortion distribution map}.
\newblock \bibinfo{journal}{\emph{IEEE Transactions on Image Processing}}
  \bibinfo{volume}{29} (\bibinfo{year}{2020}), \bibinfo{pages}{3707--3718}.
\newblock


\bibitem[Liu et~al\mbox{.}(2021b)]%
        {VST}
\bibfield{author}{\bibinfo{person}{Nian Liu}, \bibinfo{person}{Ni Zhang},
  \bibinfo{person}{Kaiyuan Wan}, \bibinfo{person}{Ling Shao}, {and}
  \bibinfo{person}{Junwei Han}.} \bibinfo{year}{2021}\natexlab{b}.
\newblock \showarticletitle{Visual Saliency Transformer}. In
  \bibinfo{booktitle}{\emph{Proceedings of the IEEE/CVF International
  Conference on Computer Vision(ICCV)}}.
\newblock


\bibitem[Liu et~al\mbox{.}(2021a)]%
        {liu2021swin}
\bibfield{author}{\bibinfo{person}{Ze Liu}, \bibinfo{person}{Yutong Lin},
  \bibinfo{person}{Yue Cao}, \bibinfo{person}{Han Hu}, \bibinfo{person}{Yixuan
  Wei}, \bibinfo{person}{Zheng Zhang}, \bibinfo{person}{Stephen Lin}, {and}
  \bibinfo{person}{Baining Guo}.} \bibinfo{year}{2021}\natexlab{a}.
\newblock \showarticletitle{Swin transformer: Hierarchical vision transformer
  using shifted windows}. In \bibinfo{booktitle}{\emph{Proceedings of the
  IEEE/CVF International Conference on Computer Vision(ICCV)}}.
\newblock


\bibitem[Ma et~al\mbox{.}(2020)]%
        {360-SSOD}
\bibfield{author}{\bibinfo{person}{Guangxiao Ma}, \bibinfo{person}{Shuai Li},
  \bibinfo{person}{Chenglizhao Chen}, \bibinfo{person}{Aimin Hao}, {and}
  \bibinfo{person}{Hong Qin}.} \bibinfo{year}{2020}\natexlab{}.
\newblock \showarticletitle{Stage-wise Salient Object Detection in
  $360^{\circ}$ Omnidirectional Image via Object-level Semantical Saliency
  Ranking}.
\newblock \bibinfo{journal}{\emph{IEEE Transactions on Visualization and
  Computer Graphics}} \bibinfo{volume}{26}, \bibinfo{number}{12}
  (\bibinfo{year}{2020}), \bibinfo{pages}{3535--3545}.
\newblock


\bibitem[Nishiyama et~al\mbox{.}(2021)]%
        {nishiyama2021360}
\bibfield{author}{\bibinfo{person}{Akito Nishiyama}, \bibinfo{person}{Satoshi
  Ikehata}, {and} \bibinfo{person}{Kiyoharu Aizawa}.}
  \bibinfo{year}{2021}\natexlab{}.
\newblock \showarticletitle{360 single image super resolution via
  distortion-aware network and distorted perspective images}. In
  \bibinfo{booktitle}{\emph{Proceedings of the IEEE International Conference on
  Image Processing (ICIP)}}.
\newblock


\bibitem[Pang et~al\mbox{.}(2022)]%
        {Pang_2022_CVPR}
\bibfield{author}{\bibinfo{person}{Youwei Pang}, \bibinfo{person}{Xiaoqi Zhao},
  \bibinfo{person}{Tian-Zhu Xiang}, \bibinfo{person}{Lihe Zhang}, {and}
  \bibinfo{person}{Huchuan Lu}.} \bibinfo{year}{2022}\natexlab{}.
\newblock \showarticletitle{Zoom in and Out: A Mixed-Scale Triplet Network for
  Camouflaged Object Detection}. In \bibinfo{booktitle}{\emph{Proceedings of
  the IEEE/CVF Conference on Computer Vision and Pattern Recognition (CVPR)}}.
\newblock


\bibitem[Paszke et~al\mbox{.}(2019)]%
        {Pytorch}
\bibfield{author}{\bibinfo{person}{Adam Paszke}, \bibinfo{person}{Sam Gross},
  \bibinfo{person}{Francisco Massa}, \bibinfo{person}{Adam Lerer},
  \bibinfo{person}{James Bradbury}, \bibinfo{person}{Gregory Chanan},
  \bibinfo{person}{Trevor Killeen}, \bibinfo{person}{Zeming Lin},
  \bibinfo{person}{Natalia Gimelshein}, \bibinfo{person}{Luca Antiga},
  {et~al\mbox{.}}} \bibinfo{year}{2019}\natexlab{}.
\newblock \showarticletitle{Pytorch: An imperative style, high-performance deep
  learning library}.
\newblock \bibinfo{journal}{\emph{Advances in neural information processing
  systems}}  \bibinfo{volume}{32} (\bibinfo{year}{2019}).
\newblock


\bibitem[Rai et~al\mbox{.}(2017)]%
        {Salient!360}
\bibfield{author}{\bibinfo{person}{Yashas Rai}, \bibinfo{person}{Jes{\'u}s
  Guti{\'e}rrez}, {and} \bibinfo{person}{Patrick Le~Callet}.}
  \bibinfo{year}{2017}\natexlab{}.
\newblock \showarticletitle{A dataset of head and eye movements for 360 degree
  images}. In \bibinfo{booktitle}{\emph{Proceedings of the 8th ACM on
  Multimedia Systems Conference}}.
\newblock


\bibitem[Shen et~al\mbox{.}(2022)]%
        {shen2022panoformer}
\bibfield{author}{\bibinfo{person}{Zhijie Shen}, \bibinfo{person}{Chunyu Lin},
  \bibinfo{person}{Kang Liao}, \bibinfo{person}{Lang Nie},
  \bibinfo{person}{Zishuo Zheng}, {and} \bibinfo{person}{Yao Zhao}.}
  \bibinfo{year}{2022}\natexlab{}.
\newblock \showarticletitle{PanoFormer: Panorama Transformer for Indoor
  $360^{\circ}$ Depth Estimation}. In \bibinfo{booktitle}{\emph{Proceedings of
  the European Conference on Computer Vision (ECCV)}}. Springer,
  \bibinfo{pages}{195--211}.
\newblock


\bibitem[Sitzmann et~al\mbox{.}(2018)]%
        {Stanford360}
\bibfield{author}{\bibinfo{person}{Vincent Sitzmann}, \bibinfo{person}{Ana
  Serrano}, \bibinfo{person}{Amy Pavel}, \bibinfo{person}{Maneesh Agrawala},
  \bibinfo{person}{Diego Gutierrez}, \bibinfo{person}{Belen Masia}, {and}
  \bibinfo{person}{Gordon Wetzstein}.} \bibinfo{year}{2018}\natexlab{}.
\newblock \showarticletitle{Saliency in VR: How do people explore virtual
  environments?}
\newblock \bibinfo{journal}{\emph{IEEE transactions on visualization and
  computer graphics}} \bibinfo{volume}{24}, \bibinfo{number}{4}
  (\bibinfo{year}{2018}), \bibinfo{pages}{1633--1642}.
\newblock


\bibitem[Su et~al\mbox{.}(2019)]%
        {su2019selectivity}
\bibfield{author}{\bibinfo{person}{Jinming Su}, \bibinfo{person}{Jia Li},
  \bibinfo{person}{Yu Zhang}, \bibinfo{person}{Changqun Xia}, {and}
  \bibinfo{person}{Yonghong Tian}.} \bibinfo{year}{2019}\natexlab{}.
\newblock \showarticletitle{Selectivity or invariance: Boundary-aware salient
  object detection}. In \bibinfo{booktitle}{\emph{Proceedings of the IEEE/CVF
  International Conference on Computer Vision(ICCV)}}.
\newblock


\bibitem[Tateno et~al\mbox{.}(2018)]%
        {tateno2018distortion}
\bibfield{author}{\bibinfo{person}{Keisuke Tateno}, \bibinfo{person}{Nassir
  Navab}, {and} \bibinfo{person}{Federico Tombari}.}
  \bibinfo{year}{2018}\natexlab{}.
\newblock \showarticletitle{Distortion-aware convolutional filters for dense
  prediction in panoramic images}. In \bibinfo{booktitle}{\emph{Proceedings of
  the European Conference on Computer Vision (ECCV)}}.
\newblock


\bibitem[Vaswani et~al\mbox{.}(2017)]%
        {transformer}
\bibfield{author}{\bibinfo{person}{Ashish Vaswani}, \bibinfo{person}{Noam
  Shazeer}, \bibinfo{person}{Niki Parmar}, \bibinfo{person}{Jakob Uszkoreit},
  \bibinfo{person}{Llion Jones}, \bibinfo{person}{Aidan~N Gomez},
  \bibinfo{person}{\L~ukasz Kaiser}, {and} \bibinfo{person}{Illia Polosukhin}.}
  \bibinfo{year}{2017}\natexlab{}.
\newblock \showarticletitle{Attention is All you Need}. In
  \bibinfo{booktitle}{\emph{Advances in Neural Information Processing
  Systems}}, Vol.~\bibinfo{volume}{30}.
\newblock


\bibitem[Wang et~al\mbox{.}(2020)]%
        {Wang_2020_CVPR}
\bibfield{author}{\bibinfo{person}{Fu-En Wang}, \bibinfo{person}{Yu-Hsuan Yeh},
  \bibinfo{person}{Min Sun}, \bibinfo{person}{Wei-Chen Chiu}, {and}
  \bibinfo{person}{Yi-Hsuan Tsai}.} \bibinfo{year}{2020}\natexlab{}.
\newblock \showarticletitle{BiFuse: Monocular 360 Depth Estimation via
  Bi-Projection Fusion}. In \bibinfo{booktitle}{\emph{Proceedings of the
  IEEE/CVF Conference on Computer Vision and Pattern Recognition (CVPR)}}.
\newblock


\bibitem[Wang et~al\mbox{.}(2017)]%
        {wang2017stagewise}
\bibfield{author}{\bibinfo{person}{Tiantian Wang}, \bibinfo{person}{Ali Borji},
  \bibinfo{person}{Lihe Zhang}, \bibinfo{person}{Pingping Zhang}, {and}
  \bibinfo{person}{Huchuan Lu}.} \bibinfo{year}{2017}\natexlab{}.
\newblock \showarticletitle{A stagewise refinement model for detecting salient
  objects in images}. In \bibinfo{booktitle}{\emph{Proceedings of the IEEE
  International Conference on Computer Vision(ICCV)}}.
\newblock


\bibitem[Wang et~al\mbox{.}(2023)]%
        {wang2023pixels}
\bibfield{author}{\bibinfo{person}{Yi Wang}, \bibinfo{person}{Ruili Wang},
  \bibinfo{person}{Xin Fan}, \bibinfo{person}{Tianzhu Wang}, {and}
  \bibinfo{person}{Xiangjian He}.} \bibinfo{year}{2023}\natexlab{}.
\newblock \showarticletitle{Pixels, Regions, and Objects: Multiple Enhancement
  for Salient Object Detection}. In \bibinfo{booktitle}{\emph{Proceedings of
  the IEEE/CVF Conference on Computer Vision and Pattern Recognition(CVPR)}}.
\newblock


\bibitem[Woo et~al\mbox{.}(2018)]%
        {woo2018cbam}
\bibfield{author}{\bibinfo{person}{Sanghyun Woo}, \bibinfo{person}{Jongchan
  Park}, \bibinfo{person}{Joon-Young Lee}, {and} \bibinfo{person}{In~So
  Kweon}.} \bibinfo{year}{2018}\natexlab{}.
\newblock \showarticletitle{Cbam: Convolutional block attention module}. In
  \bibinfo{booktitle}{\emph{Proceedings of the European Conference on Computer
  Vision (ECCV)}}.
\newblock


\bibitem[Wu et~al\mbox{.}(2022)]%
        {ODI-SOD}
\bibfield{author}{\bibinfo{person}{Junjie Wu}, \bibinfo{person}{Changqun Xia},
  \bibinfo{person}{Tianshu Yu}, {and} \bibinfo{person}{Jia Li}.}
  \bibinfo{year}{2022}\natexlab{}.
\newblock \showarticletitle{View-aware Salient Object Detection for
  $360^{\circ}$ Omnidirectional Image}.
\newblock \bibinfo{journal}{\emph{IEEE Transactions on Multimedia}}
  (\bibinfo{year}{2022}), \bibinfo{pages}{1--15}.
\newblock


\bibitem[Wu et~al\mbox{.}(2019)]%
        {wu2019cascaded}
\bibfield{author}{\bibinfo{person}{Zhe Wu}, \bibinfo{person}{Li Su}, {and}
  \bibinfo{person}{Qingming Huang}.} \bibinfo{year}{2019}\natexlab{}.
\newblock \showarticletitle{Cascaded partial decoder for fast and accurate
  salient object detection}. In \bibinfo{booktitle}{\emph{Proceedings of the
  IEEE/CVF Conference on Computer Vision and Pattern Recognition(CVPR)}}.
\newblock


\bibitem[Yoon et~al\mbox{.}(2022)]%
        {yoon2022spheresr}
\bibfield{author}{\bibinfo{person}{Youngho Yoon}, \bibinfo{person}{Inchul
  Chung}, \bibinfo{person}{Lin Wang}, {and} \bibinfo{person}{Kuk-Jin Yoon}.}
  \bibinfo{year}{2022}\natexlab{}.
\newblock \showarticletitle{SphereSR: 360deg Image Super-Resolution With
  Arbitrary Projection via Continuous Spherical Image Representation}. In
  \bibinfo{booktitle}{\emph{Proceedings of the IEEE/CVF Conference on Computer
  Vision and Pattern Recognition(CVPR)}}.
\newblock


\bibitem[Yuan et~al\mbox{.}(2021)]%
        {yuan2021tokens}
\bibfield{author}{\bibinfo{person}{Li Yuan}, \bibinfo{person}{Yunpeng Chen},
  \bibinfo{person}{Tao Wang}, \bibinfo{person}{Weihao Yu},
  \bibinfo{person}{Yujun Shi}, \bibinfo{person}{Zi-Hang Jiang},
  \bibinfo{person}{Francis~EH Tay}, \bibinfo{person}{Jiashi Feng}, {and}
  \bibinfo{person}{Shuicheng Yan}.} \bibinfo{year}{2021}\natexlab{}.
\newblock \showarticletitle{Tokens-to-token vit: Training vision transformers
  from scratch on imagenet}. In \bibinfo{booktitle}{\emph{Proceedings of the
  IEEE/CVF International Conference on Computer Vision(ICCV)}}.
\newblock


\bibitem[Zhang et~al\mbox{.}(2019)]%
        {zhang2019orientation}
\bibfield{author}{\bibinfo{person}{Chao Zhang}, \bibinfo{person}{Stephan
  Liwicki}, \bibinfo{person}{William Smith}, {and} \bibinfo{person}{Roberto
  Cipolla}.} \bibinfo{year}{2019}\natexlab{}.
\newblock \showarticletitle{Orientation-aware semantic segmentation on
  icosahedron spheres}. In \bibinfo{booktitle}{\emph{Proceedings of the
  IEEE/CVF International Conference on Computer Vision(ICCV)}}.
\newblock


\bibitem[Zhang et~al\mbox{.}(2022)]%
        {zhang2022bending}
\bibfield{author}{\bibinfo{person}{Jiaming Zhang}, \bibinfo{person}{Kailun
  Yang}, \bibinfo{person}{Chaoxiang Ma}, \bibinfo{person}{Simon Rei{\ss}},
  \bibinfo{person}{Kunyu Peng}, {and} \bibinfo{person}{Rainer Stiefelhagen}.}
  \bibinfo{year}{2022}\natexlab{}.
\newblock \showarticletitle{Bending reality: Distortion-aware transformers for
  adapting to panoramic semantic segmentation}. In
  \bibinfo{booktitle}{\emph{Proceedings of the IEEE/CVF Conference on Computer
  Vision and Pattern Recognition(CVPR)}}.
\newblock


\bibitem[Zhang et~al\mbox{.}(2020)]%
        {F-360iSOD}
\bibfield{author}{\bibinfo{person}{Yi Zhang}, \bibinfo{person}{Lu Zhang},
  \bibinfo{person}{Wassim Hamidouche}, {and} \bibinfo{person}{Olivier
  Deforges}.} \bibinfo{year}{2020}\natexlab{}.
\newblock \showarticletitle{A Fixation-Based $360^{\circ}$ Benchmark Dataset
  For Salient Object Detection}. In \bibinfo{booktitle}{\emph{Proceedings of
  the IEEE International Conference on Image Processing (ICIP)}}.
\newblock


\bibitem[Zhao et~al\mbox{.}(2019)]%
        {EGNet}
\bibfield{author}{\bibinfo{person}{Jiaxing Zhao}, \bibinfo{person}{Jiang-Jiang
  Liu}, \bibinfo{person}{Deng-Ping Fan}, \bibinfo{person}{Yang Cao},
  \bibinfo{person}{Jufeng Yang}, {and} \bibinfo{person}{Ming-Ming Cheng}.}
  \bibinfo{year}{2019}\natexlab{}.
\newblock \showarticletitle{EGNet: Edge Guidance Network for Salient Object
  Detection}. In \bibinfo{booktitle}{\emph{Proceedings of the IEEE/CVF
  International Conference on Computer Vision(ICCV)}}.
\newblock


\bibitem[Zhao et~al\mbox{.}(2020b)]%
        {zhao2020depth}
\bibfield{author}{\bibinfo{person}{Jiawei Zhao}, \bibinfo{person}{Yifan Zhao},
  \bibinfo{person}{Jia Li}, {and} \bibinfo{person}{Xiaowu Chen}.}
  \bibinfo{year}{2020}\natexlab{b}.
\newblock \showarticletitle{Is depth really necessary for salient object
  detection?}. In \bibinfo{booktitle}{\emph{Proceedings of the 28th ACM
  international conference on Multimedia}}.
\newblock


\bibitem[Zhao et~al\mbox{.}(2020a)]%
        {GateNet}
\bibfield{author}{\bibinfo{person}{Xiaoqi Zhao}, \bibinfo{person}{Youwei Pang},
  \bibinfo{person}{Lihe Zhang}, \bibinfo{person}{Huchuan Lu}, {and}
  \bibinfo{person}{Lei Zhang}.} \bibinfo{year}{2020}\natexlab{a}.
\newblock \showarticletitle{Suppress and Balance: A Simple Gated Network for
  Salient Object Detection}. In \bibinfo{booktitle}{\emph{Proceedings of the
  European Conference on Computer Vision (ECCV)}}.
\newblock


\end{thebibliography}
}

\section*{Appendix}
\subsection*{Comparison with More Baseline Methods}
For the latest 360 SOD methods, we were unable to compare them on all datasets due to some methods not being open-source yet. Therefore, we conducted comparisons on common evaluation metrics specifically on the 360-SOD dataset.
We have included two additional 360 SOD methods, \textit{i.e.,} MPFR-Net~\cite{cong2023multi} and MFFPANet~\cite{MFFPANet}, for comparison. The results are presented in Table~\ref{tab:latest_methods}. It should be noted that the results of these two methods were directly copied from their original papers. It is evident that our proposed method outperforms both MPFR-Net and MFFPANet on the 360-SOD dataset, even though MFFPANet is a multi-stage method.

In addition, we compared our method with a very recent SOD method, MENet (CVPR 2023)~\cite{wang2023pixels}. However, it should be noted that MENet is primarily designed for 2D natural images and not optimized for handling 360° data. As a result, MENet exhibited relatively lower performance compared to our method on the 360-SOD dataset.

\appendix
\setcounter{table}{0}
\renewcommand{\thetable}{A\arabic{table}}
\begin{table}[h]
\footnotesize 
\centering
\begin{tabular}{c|cccccc}
\hline
Methods & $MAE\downarrow$ & $maxF\uparrow$ & $meanF\uparrow$ & $maxE\uparrow$ & $meanE\uparrow$ & $S_m\uparrow$  \\
\hline
MPFR-Net & 0.019 & - & 0.754 & - & 0.892 & 0.842\\
\hline
MFFPANet & 0.019 & 0.723 & 0.682 & 0.905 & 0.815 & 0.788 \\
\hline
MENet & 0.0243 & 0.7129 & 0.6830 & 0.8837 & 0.8752 & 0.7963 \\
\hline
Ours & \textbf{0.0174} & \textbf{0.7928} & \textbf{0.7742} & \textbf{0.9191} & \textbf{0.9071} & \textbf{0.8493} \\
\hline
\end{tabular}
\caption{Comparison between latest SOD methods and our proposed method on the 360-SOD dataset. }
\label{tab:latest_methods}
\end{table}

\subsection*{More Experiments about Distortion-adaptive Attention Block}
In our method, we integrated the distortion adaptation module into the Decoder instead of incorporating it into the Encoder. In fact, our current designs were experimentally decided. Here we provide comparison of different variants in Table~\ref{tab:Encoder}: 

1. \textbf{Variant 1: Encoder+DM} - We move the DM module from the decoder to the encoder; 

2. \textbf{Variant 2: Encoder+DA-frozen} - We replace the Transformer Layer in the encoder with our DA block while keeping the pre-trained encoder frozen; 

3. \textbf{Variant 3: Encoder+DA-finetune} - Similar to Variant 2, the pre-trained encoder is trainable; 

4. \textbf{Variant 4: Encoder+DA-parallel} - We add the DA block to the encoder in parallel with the original Transformer Layer; 

5. \textbf{Variant 5: Baseline+DA-parallel} - We only add DA block to the encoder in parallel with the original Transformer Layer in our baseline model. 

\begin{table}[h]
\footnotesize 
\centering
\begin{tabular}{c|cccccc}
\hline
Methods & $MAE\downarrow$ & $maxF\uparrow$ & $meanF\uparrow$ & $maxE\uparrow$ & $meanE\uparrow$ & $S_m\uparrow$  \\
\hline
Variant 1 & 0.0197 & 0.7458 & 0.7308 & 0.8776 & 0.8555 & 0.8240 \\
\hline
Variant 2 & 0.0236 & 0.6745 & 0.6607 & 0.8290 & 0.8047 & 0.7769 \\
\hline
Variant 3 & 0.0207 & 0.7239 & 0.7064 & 0.8700 & 0.8490 & 0.8127 \\
\hline
Variant 4 & 0.0177 & 0.7850 & 0.7626 & 0.9028 & 0.8889 & 0.8444 \\
\hline
Variant 5 & 0.0198 & 0.7676 & 0.7362 & 0.8972 & 0.8799 & 0.8341 \\
\hline
Baseline & 0.0199 & 0.7523 & 0.7295 & 0.8878 & 0.8659 & 0.8186 \\
\hline
Ours & \textbf{0.0174} & \textbf{0.7928} & \textbf{0.7742} & \textbf{0.9191} & \textbf{0.9071} & \textbf{0.8493} \\
\hline
\end{tabular}
\caption{Comparison between different model variants and the proposed model on the 360-SOD dataset.}
\label{tab:Encoder}
\end{table}

The results show that the current configuration of our model outperforms all other variants. We speculate the main reason behind these results is that the newly added components may degrade the feature extraction ability of the pre-trained encoder especially considering the limited scale of the training data.

\subsection*{More Experiments about Relation Matrix}
In our proposed model, we proposed a relation matrix to utilize the spatial prior observed in 360° images. 
The work~\cite{nishiyama2021360} also introduced a prior map for 360° image super resolution task. However, in their approach, they directly concatenate the prior map with the input image. 
Therefore, we conducted a comparison between our proposed relation matrix and the approach used in \cite{nishiyama2021360}, \textit{i.e.,} concatenating the prior with the input image.  The results are shown in Table~\ref{tab:concat_RM}. It is clear to see that our method performs much better than this alternative approach, highlighting the effectiveness of the way we utilize the spatial prior.

\begin{table}[h]
\footnotesize
\centering
\begin{tabular}{c|cccccc}
\hline
Methods & $MAE\downarrow$ & $maxF\uparrow$ & $meanF\uparrow$ & $maxE\uparrow$ & $meanE\uparrow$ & $S_m\uparrow$  \\
\hline
Concat & 0.0187 & 0.7455 & 0.7298 & 0.8757 & 0.8607 & 0.8285\\
\hline
Ours & \textbf{0.0174} & \textbf{0.7928} & \textbf{0.7742} & \textbf{0.9191} & \textbf{0.9071} & \textbf{0.8493} \\
\hline
\end{tabular}
\caption{Comparison of results using different strategies for utilizing the spatial prior. The results are obtained on the 360-SOD dataset.}
\label{tab:concat_RM}
\end{table}

\end{document}